\documentclass[11pt]{article}
\usepackage{coling2016}
\usepackage{times}
\usepackage{url}
\usepackage{latexsym}
\usepackage{array}
\usepackage{latexsym}
\usepackage{footmisc}
\usepackage{comment}
\usepackage{amsmath}
\usepackage{subcaption}
\usepackage{xcolor}
\usepackage{graphicx}
\usepackage{hyperref}
\usepackage{epstopdf}
\usepackage{multirow}
\usepackage{tabularx}
\usepackage{enumitem}
\usepackage[]{algorithm2e}
\usepackage{color}

\title{A Two-Phase Approach Towards Identifying Argument Structure in Natural Language}

\author{Arkanath Pathak \\
  Deptt. Computer Sc \& Engg.\\
  IIT Kharagpur \\  
  \\\And
  Pawan Goyal \\
  Deptt. Computer Sc \& Engg. \\
  IIT Kharagpur \\
  {\tt pathak.arkanath@gmail.com \{pawang@cse, plaban@cet\}.iitkgp.ernet.in} \\\And
  Plaban Bhowmick \\
  Centre for Educational Tech. \\
  IIT Kharagpur \\
  }

\date{}

\begin{document}

\maketitle

\begin{abstract}
We propose a new approach for extracting argument structure from natural language texts that contain an underlying argument. Our approach comprises of two phases: Score Assignment and Structure Prediction. The Score Assignment phase trains models to classify relations between argument units (Support, Attack or Neutral). To that end, different training strategies have been explored. We identify different linguistic and lexical features for training the classifiers. Through ablation study, we observe that our novel use of word-embedding features is most effective for this task. The Structure Prediction phase makes use of the scores from the Score Assignment phase to arrive at the optimal structure. We perform experiments on three argumentation datasets, namely, AraucariaDB, Debatepedia and Wikipedia. We also propose two baselines and observe that the proposed approach outperforms baseline systems for the final task of Structure Prediction.
\end{abstract}

\section{Introduction}
\blfootnote{This work is licensed under a Creative Commons Attribution 4.0 International License. License details: \url{http://creativecommons.org/licenses/by/4.0/}}
The problem of argumentation mining concerns the identification of argument structures in a text. The argument structure is typically represented as a directed graph with textual propositions as nodes and both Support and Attack relations as edges between the propositions. In their influential work, \newcite{mochales2011argumentation} have discussed this problem in detail together with the relevant definitions, frameworks, and terminologies. They define the argumentation structure as consisting of various ``arguments", forming a tree structure, where each argument consists of a single conclusion and one or more premise(s). Another widely used framework is the Freeman theory of argumentation structures \cite{freeman1991dialectics,freeman2011argument}, which treats an argument as a set of proponent or opponent propositions for a central claim. In the present study, we follow the framework used in \cite{mochales2011argumentation}.

The full task of argumentation mining can be divided into four subtasks \cite{mochales2011argumentation}: 
\begin{enumerate}
\item{\textbf{Segmentation}: Splitting the text into propositions.}
\item{\textbf{Detection}: Identifying the argumentative propositions.}
\item{\textbf{Classification}: Classifying the argumentative propositions into pre-defined classes (e.g. premise or conclusion in the Mochales and Moens' framework and proponent or opponent in the case of Freeman's framework).}
\item{\textbf{Structure Prediction}: Building the structure by identifying the relations (the edges in the argumentative graph structure) between the propositions.}
\end{enumerate}
Our work jointly tackles the Classification and Structure Prediction subtasks using a unified approach. Little work has been done as far as Structure Prediction is concerned. We are familiar with only two approaches that can be compared to our work. \newcite{lawrence2014mining} proposed to form bidirectional edges between propositions in their work on $19^{th}$ century philosophical texts. They used Euclidean distance metric between topic measures derived from a generated topic model for the text to be studied. They achieved a raw accuracy of 33\% for linking the edges. However, they do not form directed edges between the argumentation units, which is essential in case of arguments. \newcite{peldszus2015joint} jointly predict different aspects of the argument structure. Recent works \newcite{persing2016end,stab2016parsing} identified argument relations in the student essays. Discourse features (positional features) have been used in these studies However, absence of these features in argument graph dataset like AraucariaDB (used in our study) makes the relation classification task challenging. 

This paper makes the following contributions. i). We propose a data-driven approach for identifying argument structure in natural language text. ii). We present a detailed study over various linguistic, structural and semantic features properties involved in the argument relation classification task. iii). Finally, we propose a metric for evaluating performance of the structure prediction task.
\section{Problem Formulation} \label{Goal}
\begin{figure}[!thb]
\vspace{-0.3cm}
\centering{\includegraphics[width=0.45\linewidth]{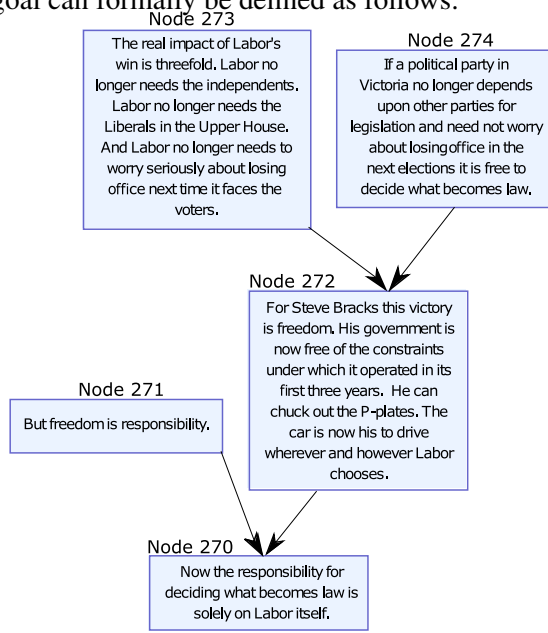}}
\caption{Sample argument (Argument No. 9) from the AraucariaDB.}
\label{argument9}
\vspace{-0.3cm}
\end{figure}

\noindent We have used the AraucariaDB \cite{reed2004araucaria} dataset\footnote{The AraucariaDB dataset can be downloaded and visualized with the AIFdb \cite{lawrence2012aifdb} framework at \href{http://www.arg.dundee.ac.uk/aif-corpora/araucaria}{http://www.arg.dundee.ac.uk/aif-corpora/araucaria}.} to discuss the problem at hand. This corpus consists of 661 argument structures, collected and analysed as a part of a project at the University of Dundee (UK). We found AraucariaDB to be one of the most suitable argumentation datasets, primarily because it is formed from natural language resources like newspapers and magazines. Fig. \ref{argument9} shows an argument structure from AraucariaDB. The edges in the tree represent a Support relation. For instance, Node 271 and Node 272 are the children of Node 270. Therefore, $271 \rightarrow 270$ as well as $272 \rightarrow 270$ are Support relations. For a given Support relation $n_1 \rightarrow n_2$, we call the node $n_1$ as the \textit{Text} node and $n_2$ as the \textit{Hypothesis} node. 
\vspace{-0.2cm}
Our goal can formally be defined as follows:

\textit{For a given set of propositions, with an underlying argument structure connecting these propositions, identify the argument structure, that is structurally close to the corresponding gold standard argument.}

A measure for structure similarity (which we call $SimScore$) is described in Section \ref{PredictionResults}.
\section{Proposed Approach} \label{Approach}

We model an argument structure as a graph, where each node represents a proposition. Given a set of nodes $N$ for an argument, our approach can be divided into two subtasks:\footnote{We assume only Support and Neutral relations between arguments, and the final structure to be a tree. In Section~\ref{AttackExperiments}, we extend this model to include Attack relations as well as linear structures.}

\noindent{\textbf{Score Assignment} : Assign scores $s_{n_1,n_2}\in[0,1]$ for every pair of nodes $n_1,n_2 \in N$, $n_1 \neq n_2$,. These scores represent the degree of Support relation present in the hypothetical edge connecting $n_1$ and $n_2$.}

\noindent{\textbf{Structure Prediction} : Choose the tree $T^*$ with the maximum sum of edge scores, i.e.,\\
$T^* = \underset{T} {\mathrm{argmax}} ~\sum\limits_{(n_1,n_2)\in E(T)} s_{n_1,n_2}$

where $T$ can be any tree with the set of nodes $N$ and $E(T)$ denotes the set of edges in $T$.}
We use the confidence measures provided by machine learning classifiers as the edge scores in Score Assignment. Specifically, we use binary classification with the classes being \textit{Support} and \textit{Neutral}. The classifier takes as input an ordered pair ($n_1$,$n_2$), where $n_1$ and $n_2$ are the text nodes.

For Stucture Prediction, our implementation essentially iterates over all possible tree structures (for the given set of nodes) recursively to choose the best tree. A call to the recursive function will already have the tree structure formed upto the last level. The function iterates on all possible sets of nodes (subset of the set of remaining nodes) for the next level. The parent of each node in the next level can be identified (from the last level) as the node which gives rise to the best attachment score. The complexity of our implementation is exponential in the number of nodes in the argument. Therefore, we limit the experiments usually to arguments with 10-15 nodes.

\section{Classifier Features} \label{Features}
\vspace{-0.2cm}
In this section, we describe the set of features chosen for the Score Assignment subtask. The task of Score Assignment is similar to Recognizing Textual Entailment (RTE) or the detection of Natural Language Inference (NLI). However, the problem is still considerably different in the case of arguments since the types of arguments can be much more complex \cite{walton2007visualization}. To make this more evident, we have formed a baseline for our experiments (Section \ref{Results}) where we use a state-of-the-art RTE tool instead for the Score Assignment subtask. We experimented with the features frequently used in NLI, RTE and similar tasks. \newcite{maccartney2009natural} discusses the features used for the NLI task in detail. However, some of the frequently used features like POS n-grams, the length of propositions, POS of the main verb, etc.,  are not included in the set of features because they showed insignificant effect on the overall performance in our experiments. We suspect this is due to the fact that many attributes of these features are already captured in the features we have chosen.

\noindent\textbf{Discourse Markers:} These are the words that are indicative of argumentative discourse. Discourse markers have persistently been used for both RTE and NLI tasks. \newcite{eckle2015role} have also discussed the role of discourse markers in the context of argumentation mining. However, we observed that the presence of such words are rare in the AraucariaDB dataset. We have used i). counts of the following words in Text: 
     \textit{as}, \textit{or}, \textit{and}, \textit{roughly}, \textit{then}, \textit{since},
   and ii). counts of the following words in Hypothesis: 
    \textit{therefore}, \textit{however}, \textit{though}, \textit{but}, \textit{quite}.
This feature set gives rise to 11 (6$+$5) features.

\noindent\textbf{Modal Features:} These are similar to discourse markers but they do not inherently belong to either one of Text or Hypothesis. Therefore, we take the counts of these as features for both Text as well as Hypothesis inputs. We have used the following as the modal words:
     \textit{can}, \textit{could}, \textit{may}, \textit{might}, \textit{must}, \textit{will}, \textit{would}, \textit{should}. 
Modal feature set, therefore, gives rise to 16 (8$\times$2) features.

\noindent\textbf{Longest Common Phrase:} The number of words in the longest phrase present in both Text and Hypothesis is considered as a single feature.

\noindent\textbf{Common Wikipedia Entities:} In many cases, a specific argument usually revolves around some conceptual entities. For example, Argument 9 (Fig.\ref{argument9}) involves the entities \textit{Steve Bracks}, \textit{Labor}, etc. We have used TAGME \cite{ferragina2010fast} to annotate a text with Wikipedia entities. After we have the annotations as a vector for both Text and Hypothesis, we take the inner product of the resulting vectors as a single feature.

\noindent\textbf{Word N-grams:} Word n-grams are used very frequently as features for NLP tasks. We have used the set of unigrams and bigrams filtered by relative likelihood of presence in Text or Hypothesis nodes in the training data. For instance, the n-grams with higher values of $\frac{p(ngram|Text)}{p(ngram|Hypothesis)}$ will be chosen from Text nodes. The filtering is performed using a constant threshold parameter of the likelihood. We have set the threshold parameter as 3 for all of our experiments. Since we have performed Cross Validation, the number of features for this category will be different for each fold. The mean count for unigrams was 115.4 and that for bigrams was 251.8. Hence, an average of 734.4 (115.4$\times$2 + 251.8$\times$2) n-gram features were used over the 5 folds.

\noindent\textbf{Word Vector Embeddings:} Word vectors capture a variety of helpful information in the context of natural language. We have directly used the 300-dimensional vectors trained on part of the Google News dataset \cite{mikolov2013distributed} \footnote{The pre-trained word vectors for Google News dataset are freely available at \href{https://code.google.com/archive/p/word2vec/}{https://code.google.com/archive/p/word2vec/}.}. We have used the sum of word vectors over words present in Text node to form a feature vector. To generate another feature vector, a similar process is repeated for Hypothesis node. These vectors are concatenated to give rise to 600 features for an input pair.  
Using word vectors as features can help with various attributes inherent to Support edges. First, a simple similarity measure can be the difference of the two sum vectors, which can be well captured by using classifiers like linear SVM. Secondly, word vectors trained over an external dataset like Google News can provide the knowledge base for language not present in training set, which is very likely in case of argumentation mining since the arguments are expected to be unrelated. Lastly, since word vectors are based on contextual information, they can infer Support relation from similar contexts in training data.

\section{Experiments} \label{Results}
\vspace{-0.2cm}
All of the experiments in this section are performed on the AraucariaDB arguments. The experiments are performed in a 5-fold cross-validation framework and the mean scores are reported. The folds are over the set of arguments rather than the pairs of text nodes in order to maintain contextual independence between the folds. A subset of AraucariaDB arguments have been used in the following experiments\footnote{Due to exponential order complexity of Structure Prediction algorithm, we have selected arguments of size 10 nodes or less. Furthermore, arguments involving relations other than Support are ignored.}.

\begin{table}[!thb]
\centering
\scalebox{0.8}{
\begin{tabular}{|c|c|c|c|}
\hline
{\bf Measure} & {\begin{tabular}[x]{@{}c@{}} \bf type-1\\\bf SVM\end{tabular}} & {\begin{tabular}[x]{@{}c@{}} \bf type-2\\\bf SVM\end{tabular}} & {\begin{tabular}[x]{@{}c@{}} \bf type-2\\\bf MLP\end{tabular}} \\\hline
$confidence_S$ 			& 0.691 	& 0.643 	& 0.678	\\
$confidence_N$ 			& 0.425 	& 0.356 	& 0.306	\\
$recall_S$ 			& 0.759 	& 0.677 	& 0.677	\\
$recall_N$ 			& 0.532 	& 0.681 	& 0.692	\\
$precision_S$ 			& 0.193 	& 0.68 		& 0.688	\\
$precision_N$ 			& 0.937 	& 0.678 	& 0.681	\\
$accuracy$ 					& 0.561 	& 0.679 	& 0.684	\\ \hline
\end{tabular}
}
\caption{Classifier Performance: mean values are reported for Support ($S$) and Neutral ($N$) relations.}\label{tab:classifierresults}
\vspace{-0.7cm}
\end{table}

\subsection{Classifier Performance (Score Assignment)} \label{ClassifierResults}
\vspace{-0.15cm}
Support edges present in the input argument structures are directly taken as Support pair examples for the classifier. However, the generation of Neutral pairs is not so straightforward. To that end, we have experimented with two kinds of frameworks. Please note that the selection of training framework does not affect the ultimate goal of structure prediction. The framework used for training the classifiers is only limited to the Score Assignment phase.

The first framework, which we call the \textit{type-1} framework, considers all the pairs ($n_1$,$n_2$) as Neutral such that $n_1$ and $n_2$ are text nodes belonging to the same argument and $n_1$ $\rightarrow$ $n_2$ is not a Support relation. This, however, gives rise to a huge imbalance between the Support and Neutral examples. Many classifiers fail to perform well in such imbalance. Nonetheless, this issue can be resolved in classifiers like linear SVM, by assigning class weights inversely proportional to class frequencies \cite{king2001logistic} in the input data. We have followed this framework for type-1 SVM. Another way to resolve the problem of imbalance is to down-sample the Neutral relations randomly. However, random sampling did not perform well in our experiments, and thus, we posit that a random subset might not be a good training sample.

To counter this, we devised the \textit{type-2} framework which only considers those pairs ($n_1$,$n_2$) as Neutral for which $n_2$ $\rightarrow$ $n_1$ is a Support relation. This gives a perfectly balanced input dataset with one Neutral example corresponding to each Support example. Thereby, making it suitable for machine learning classifiers. It is difficult to compare the information captured by the two frameworks. While type-1 might seem to capture more information than type-2, type-1 is also prone to more noise since the data is larger as compared to type-2.

A Multi-layer Perceptron (MLP) classifier using the type-2 framework performed better than type-1 and type-2 SVM implementations for arguments with 3 nodes in our experiments. The network is made up of 3 hidden layers with 200 neurons for each layer.

Table \ref{tab:classifierresults} summarizes the results for the three classifiers. We have shown 7 performance measures for each classifier. Specifically, we present the mean of the confidence values provided by the classifier for each class label, which is used directly in the Structure Prediction phase. We have scaled the confidence measure\footnote{We have used the SVM and Multi-layer Perceptron classifier implementations provided by the open source library scikit-learn \cite{scikit-learn}. For the confidence measure, we have used the decision function in the case of SVM and the predicted probability in the case of MLP. Class imbalance was handled using `balanced' weighting of classes.} linearly between 0 and 1 before using it as scores for Structure Prediction. A mean confidence of 1, therefore, will be the perfect score for Support pairs. Similarly, a mean confidence of 0 will be the perfect score for Neutral pairs. We can observe that each classifier outperforms the others for at least some metric. One can observe that type-2 classifiers perform better in predicting Neutral pairs. The confidence measure is lower than type-1 and the recall is higher as well. However, the precision for Neutral is better for type-1 SVM because of the data imbalance. Type-2 MLP gave the best accuracy in our experiments. 

\subsection{Structure Prediction Performance} \label{PredictionResults}
Since the arguments are complex in nature, our approach (Section \ref{Approach}) often fails to predict the entire structure. To counter this, we formulate a measure to evaluate the similarity between the predicted tree and the input tree. The measure, $SimScore$, is defined as:
\begin{align*}
SimScore(T_1, T_2) = \frac{|E(T_1) \cap E(T_2)|}{|E(T_1)|}
\vspace{-0.2cm}
\end{align*}
where $T_1$ and $T_2$ have the same set of nodes and $E(T)$ is the set of edges for a tree T. This measure quite intuitively captures the fraction of edges common to both trees. Since the set of nodes are the same, this measure turns out to be directly related to measures like the graph edit distance \cite{sanfeliu1983distance}.

Since our problem formulation is new, it is difficult to compare the results with existing literature for Structure Prediction in argumentation mining. However, we compare our performance to two baselines. The first baseline, \textbf{Random}, is a baseline which randomly chooses any tree structure over the given set of nodes. It can be shown that the expected value of the $SimScore(T_i, T)$ for a given tree $T$ is equal to $1/n$ where $n$ is the number of nodes in $T$. For the second baseline, \textbf{EDITS}, we use the state-of-the-art RTE software package EDITS \cite{kouylekov2010open}, instead of the classifiers we proposed, for scoring the edges. However, the metric for scoring structures remains the same as the sum of edge scores. In this case, the entailment relation corresponds to Support relation. EDITS has also been used previously by \newcite{cabrio2012combining} in the context of argumentation mining.

\vspace{-0.3cm}
\begin{table*}[!thb]
\centering
\scalebox{0.8}{
\begin{tabular}{|c|c|c|c|c|c|c|}
\cline{1-7}
\multirow{2}{*}{\bf Nodes}	& \multirow{2}{*}{\bf Arguments} & \multicolumn{5}{c|}{$SimScore$} \\
\cline{3-7}
& & {\bf type-1 SVM} & {\bf type-2 SVM} & {\bf type-2 MLP} & {\bf EDITS} & {\bf Random} \\
\cline{1-7}
2 	& 10 	& \textbf{0.9} 	& 0.8 	& 0.8 	&  0.7 		& 0.5		\\
3 	& 187 	& 0.564 & 0.566 & \textbf{0.625} &  0.363 	& 0.333	\\
4 	& 85 	& 0.529 & \textbf{0.552} & 0.482 &  0.250 	& 0.250		\\
5 	& 62 	& \textbf{0.446} & 0.399 & 0.435 &  0.231 	& 0.2		\\
6 	& 72 	& \textbf{0.363} & 0.341 & 0.322 &  0.263 	& 0.166		\\
7 	& 58 	& \textbf{0.369} & 0.323 & 0.309 &  0.231 	& 0.142		\\
8 	& 41 	& \textbf{0.230} & 0.19 	& 0.199 &  0.205 	& 0.125		\\
9 	& 19		& \textbf{0.351} & 0.28 	& 0.222 &  0.265 	& 0.111		\\
10 	& 23 	& \textbf{0.217} & 0.188 & 0.115 &  0.212 	& 0.1		\\
Any	& 557 	& \textbf{0.459} & 0.442 & 0.447 &  0.289 	& 0.234		\\
\cline{1-7}
\end{tabular}
}
\caption{Structure Prediction Performance: Mean of $SimScore$ for the arguments grouped by the number of nodes in the argument. EDITS and Random are baselines whereas type-1 SVM, type-2 SVM and type-2 MLP are proposed approaches.}\label{tab:finalresults}
\vspace{-0.4cm}
\end{table*}

In Table \ref{tab:finalresults}, we compare the mean value of $SimScore$ for each classifier. We have further categorized the results based on the number of nodes present in the argument. As evident by the Random baseline, it is expected that the performance will degrade as the number of nodes increase. We can observe that all the three classifiers outperform the EDITS and Random baselines by a considerable factor. For the arguments with 3 nodes, type-2 MLP outperforms type-1 SVM and type-2 SVM. However, for arguments with higher number of nodes, type-1 SVM performs the best. The results reported are statistically significant with a p-value of 0.00198 after performing a two-tailed paired t-test between the type-1 SVM and the EDITS baseline.

\subsection{Ablation Study} \label{AblationStudy}
To test the efficacy of each individual feature/feature group, we have performed a leave-one-out ablation test. In the second column of Table \ref{tab:ablationstudy}, we report the \% decrease in the mean SimScore (for any number of nodes) when the type-1 SVM classifier is used. Following observations are made from this study.
\begin{itemize}
\item Discourse markers and modal features are observed to be the least effective feature groups.
\item Word vectors trained on an external knowledge base are highly effective.
\end{itemize}
\vspace{-0.3cm}
\begin{table*}[!thb]
\vspace{-0.3cm}
\centering
\scalebox{0.8}{
\begin{tabular}{|c|c|c|}
\cline{1-3}
\multirow{2}{*}{\bf Feature Set} & \multicolumn{2}{c|}{\% \bf decrease in \bf $SimScore$} \\
\cline{2-3} & {\bf With Word Vectors} & {\bf Without Word Vectors} \\
\cline{1-3}
Discourse Markers	& 0.09\% & 0.21\% \\
Modal Features		& 0.27\% & 0.37\% \\
Wikipedia Similarity	& 0.59\% & 0.91\% \\
Word N-grams			& 1.56\% & \textbf{21.14\%} \\
Word Vectors			& \textbf{11.4\%} & - \\
Longest Common Phrase	& 1.02\% & 1.22\% \\
\cline{1-3}
\end{tabular}
}
\caption{Ablation Study: The decrease in Structure Prediction performance due to the removal of each kind of feature. The second column corresponds to the experiment with word vectors feature set. The third column corresponds to the experiment in the absence of word vectors feature set.}\label{tab:ablationstudy}
\vspace{-0.3cm}
\end{table*}
We conjecture that word vectors encode much more information than n-grams and other linguistic features for Score Assignment. To support this conjecture, we perform another ablation test to judge the effectiveness of other features in absence of word vector feature group. The results are reported in the third column of Table \ref{tab:ablationstudy}. We observe that word n-grams are now influential with a decrease of 21.14\%, which was not the case when word vectors were present. Therefore, we can deduce that word vectors were able to capture word n-grams to a great extent. However, discourse markers and modal features are still not very influential. Discourse markers assume 3.21\% in the set of terms in AraucariaDB while the modal features are present with 1.89\%.  We hypothesize the ineffectiveness of discourse markers and modal features to their rarity in AraucariaDB. 
\vspace{-0.1cm}
\section{Arguments with Attack relations} \label{AttackExperiments}
\vspace{-0.1cm}
Till now, we have only considered arguments which solely include Support relations. A natural extension of this approach is to support the arguments which include both Support and Attack relations. In an Attack relation, $A \rightarrow B$, statement in node $A$ is used to contradict the statement of node $B$. We could not find enough argumentation datasets which include attack relations in a significant proportion. We have used two datasets, namely, \emph{Debatepedia} and \emph{Wikipedia} from NoDE \cite{cabrio2014node}\footnote{These datasets are described in detail, and are freely available for download, at \href{http://www-sop.inria.fr/NoDE/NoDE-xml.html}{http://www-sop.inria.fr/NoDE/NoDE-xml.html}.}, a benchmark of natural argument. Although these datasets are pretty small for a machine learning approach, our approach still performs reasonably well for these datasets.
\begin{figure}[h]
\begin{subfigure}{0.6\textwidth}
\includegraphics[width=\linewidth]{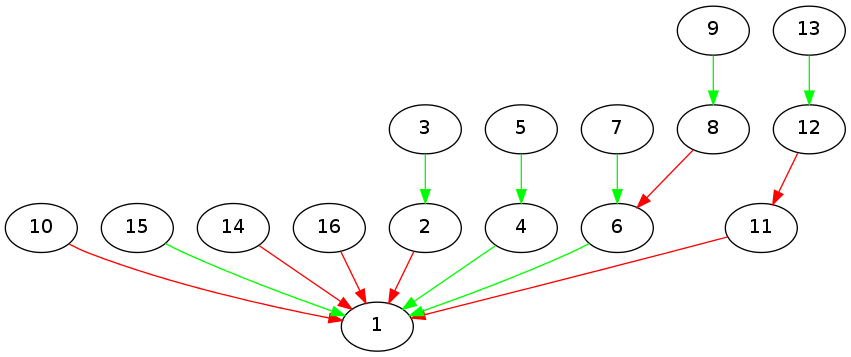}
\caption{An example debate structure from Debatepedia for the topic of Violent Games.}
\label{Debatepedia}
\end{subfigure}
\hspace*{\fill}
\begin{subfigure}{0.35\textwidth}
\centering{\includegraphics[width=0.25\linewidth]{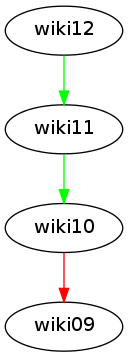}}
\caption{An example argument from the Wikipedia dataset.}
\label{wikipedia}
\end{subfigure}
\caption{Argument structures with Attack relations. Green edges indicate Support relation whereas red edges indicate Attack relation.}
\vspace{-0.6cm}
\end{figure}
The first dataset, \textit{Debatepedia}, consists of data extracted from online debate platforms (debatepedia.org and procon.org). This dataset consists of 260 (140 Support, 120 Attack) relations. Each debate is formed by the responses for a given topic. Fig. \ref{Debatepedia} shows an example debate from the dataset for the topic of ``Violent Games". We will treat such a structure for a given topic as an argument. There are 22 such topics in the dataset. We ignore one topic: ``Ground Zero Mosque", because it does not follow a tree structure. The second dataset is built on the Wikipedia revision history over a four-year period, focusing on the top five most revised articles. The Wikipedia dataset consists of 452 pairs (215 Support, 237 Attack). We consider the structure formed by the revision history to be an argument. Therefore, each argument will be a linear graph, as shown in Fig. \ref{wikipedia}. To extend our approach for decoding the best possible tree, we need to modify the algorithm to accommodate the fact that the classifiers are no longer binary.

There are now three possible relations for an ordered pair of nodes (a,b): \textbf{Support}, \textbf{Attack} and \textbf{Neutral}. In the earlier case (AraucariaDB), when we assumed the sole presence of Support relations, distinction between Support and Neutral also included the effect of features that are not related to directional inference, e.g., common Wikipedia entities. However, those features are now common to both Support and Attack relations. To account for these complications, we consider two approaches:

\textbf{Two-Step Approach}: In the first step, a classifier marks the pairs which have either Support or Attack edge. We call it the \textbf{Detection} classifier. In a similar fashion, this classifier can also be type-1 or type-2. In the second step, an independent classifier resolves those edges into Support or Attack. We call the second classifier the \textbf{Resolver}. In the Structure Prediction phase, the Detection classifier will decide the best tree structure and the Resolver classifier will then resolve the edges into either Support or Attack.

\textbf{Single-Step Approach}: A multi-class classifier resolves all the three relations: Support, Attack or Neutral. Neutral edges are generated using the type-1 framework: any possible edge formed by nodes within the argument which is not an existing Support or Attack edge. In the Structure Prediction phase, the tree structure with the maximum sum of edge scores is chosen. However, the edge score will now be $confidence_S - confidence_N$ instead of $confidence_S$ for each Support edge, where $confidence_S$ and $confidence_N$ are the confidence scores provided by the Single-Step classifier for the Support and Neutral labels, respectively. Similarly, an edge score of $confidence_A - confidence_N$ will be assigned to each Attack edge, where $confidence_A$ is the confidence score assigned by the Single-Step classifier for the Attack label. The $confidence_N$ is deducted due to the missing Neutral edge if a Support edge is chosen. It is evident that assigning $confidence_S - confidence_N$ resolves to assigning $confidence_S$ in Structure Prediction when there are no Attack labels.

For experiments on these approaches, two types of features were included in addition to the features described in Section \ref{Approach}:

\noindent \textbf{1. Negation Discourse Markers}: These markers try to capture contrast or negation sentiments in a sentence. Examples of such markers include: \textit{can't}, \textit{never}, etc. This feature set improved the Single-Step Classifier accuracy by 1.3\%.

\noindent \textbf{2. Negation/Contrast Relation Indicators}: Features in this category intend to capture negation or contrast relations present in an ordered pair of sentences. We have followed the approaches proposed in \cite{harabagiu2006negation}. This feature set improved the Single-Step Classifier accuracy by 6.4\%.

\begin{table}[!thb]
\vspace{-0.4cm}
\centering
\scalebox{0.8}{
\begin{tabular}{|c|c|c|}
\hline
\textbf{Classifier} & \textbf{Debatepedia} & \textbf{Wikipedia} \\\hline
Detection (type-1) 			& 0.804 	& 0.535	\\
Detection (type-2) 			& 0.906 	& 0.553 	\\
Resolver 			& 0.665 	& 0.719 \\
Two-Step 			& 0.560 	& 0.493 \\
Single-Step			& 0.761 	& 0.453 \\ \hline
\end{tabular}
}
\caption{Classifier Performance for datasets with Attack relations.}\label{tab:AttackExperiments}
\vspace{-0.4cm}
\end{table}

\begin{table}[!thb]
\centering
\scalebox{0.8}{
\begin{tabular}{|c|c|c|c|c|}
\hline
\textbf{Nodes} & \textbf{Arguments} & \textbf{T-S-1} & \textbf{T-S} & \textbf{S-S} \\\hline
7	&	2	&	0.83		&	0.666	&	0\\
8	&	1	&	0.57	1	&	0.428	&	0\\
9	&	1	&	0.875	&	0.5		&	0.25\\
10	&	2	&	0.721	&	0.385	&	0.055\\
11	&	4	&	0.325	&	0.225	&	0.05\\
12	&	2	&	0.545	&	0.409	&	0\\
13	&	2	&	0.541	&	0.333	&	0\\
Any	&	14	&	0.573	&	0.387	&	0.04\\ \hline
\end{tabular}
}
\caption{Mean SimScore for Debatepedia. \textit{T-S-1}: Step 1 of the Two-Step Approach \textit{T-S}: Two-Step Approach \textit{S-S}: Single-Step Approach}\label{tab:SimScoreDebatepedia}
\vspace{-0.5cm}
\end{table}

In this section, we follow a leave-one-out Cross Validation framework due to the small size of the datasets. Table \ref{tab:AttackExperiments} reports the mean classification accuracies for each classifier for the two datasets. We can see that type-2 framework performs better than type-1 for the Detection classifier. The Two-Step classifier combines the Detection (type-2) and the Resolver classification labels. These results imply that a Single-Step classification approach performs better than Two-Step for the Debatepedia dataset\footnote{Arguments having size 13 or less are used in this experiment.}. However, we shall see in Table \ref{tab:SimScoreDebatepedia} that the Single-Step approach performs poorly in the Structure Prediction phase. Table \ref{tab:SimScoreDebatepedia} reports the mean SimScore after the Structure Prediction for the Debatepedia dataset. Similar to Table \ref{tab:finalresults}, these results are additionally filtered by the number of nodes in the argument.
The third column \textbf{T-S-1}, reports performance of the Two-Step approach before the edges are resolved into Support or Attack, i.e. there is no distinction between Support and Attack edges. This is similar to the experiments we performed in Section \ref{PredictionResults}. The fourth column \textbf{T-S}, reports the overall performance of the Two-Step approach. The fifth column \textbf{S-S}, reports the performance of the Single-Step approach. We can see that Single-Step performs poorly as compared to Two-Step approach by a large margin. 
\begin{table}[!thb]
\vspace{-0.3cm}
\centering
\begin{tabular}{|c|c|c|c|c|c|}
\hline
\textbf{Nodes} & \textbf{Arg.} & \textbf{T-S-1} & \textbf{T-S} & \textbf{T-S-WL} & \textbf{S-S} \\\hline
2	&	142	&	0.507	&	0.366	&	0.366	&	0.274\\
3	&	103	&	0.441	&	0.305	&	0.262	&	0.203\\
4	&	34	&	0.254	&	0.156	&	0.176	&	0.098\\
Any	&	279	&	0.452	&	0.318	&	0.304	&	0.227\\ \hline
\end{tabular}
\caption{Mean SimScore for Wikipedia. \textit{T-S-WL}: Two-Step Approach without any restriction for linear structures. Rest of the abbreviations as per Table \ref{tab:SimScoreDebatepedia}.}\label{tab:SimScoreWikipedia}
\vspace{-0.4cm}
\end{table}
Table \ref{tab:SimScoreWikipedia} reports the mean SimScore for the Wikipedia dataset. Here we imposed an additional restriction for the structures to be linear. However, in the fifth column \textbf{T-S-WL}, we report the SimScore following the Two-Step approach without any restriction. We observe that the Single-Step approach performs relatively better for the Wikipedia dataset as compared to the Debatepedia dataset. We think this is due to the bigger arguments in Debatepedia. 

\vspace{-0.2cm}
\section{Conclusion}
\vspace{-0.2cm}
In this paper, we introduced a two-phase approach towards identification of argument structure in natural language text. The first phase involves building models for classifying text-hypothesis pairs into argument relations. The second phase makes use of the classifier confidence scores to construct the argument structure. We have proposed different training models to train the argument relation classifier. With the help of ablation study, we showed that our novel use of word vectors trained on an external corpus can be a crucial feature for such tasks, contributing as much as 11.4\% towards the performance. For the final goal of Structure Prediction, our approach predicted almost twice as many correct edges as with the random baseline. We showed that the proposed approach can be extended to arguments containing Attack relations as well, where our experiments predicted an average of 38\% edges correctly for Debatepedia dataset. 

\makeatletter
\renewcommand\@biblabel[1]{}
\makeatother

\bibliographystyle{acl}
\bibliography{ref}

\end{document}